\documentclass[11pt,a4paper]{article}
\usepackage[hyperref]{acl2020}

\usepackage{times}
\usepackage{latexsym}

\usepackage{microtype}
\usepackage{booktabs}
\usepackage{makecell}
\usepackage{tabularx}
\usepackage{array}
\newcolumntype{C}{>{\centering\arraybackslash}p{4.7em}}
\usepackage{multirow}
\usepackage{colortbl}
\usepackage{amsmath}
\usepackage{amsfonts}
\usepackage{graphicx}
\usepackage{subcaption}
\usepackage{url}

\definecolor{Gray}{gray}{0.95}

\aclfinalcopy 

\title{Specializing Unsupervised Pretraining Models \\ for Word-Level Semantic Similarity}

\author{Anne Lauscher\textsuperscript{1},  Ivan Vuli\'{c}\textsuperscript{2}, Edoardo Maria Ponti\textsuperscript{2}, Anna Korhonen\textsuperscript{2}, and Goran Glava\v{s}\textsuperscript{1} \vspace{1em} \\
  \textsuperscript{1}Data and Web Science Group, University of Mannheim, Germany \\
  \textsuperscript{2}Language Technology Lab, University of Cambridge, UK \hspace{2mm} \\
  \textsuperscript{1}{\tt \{anne,goran\}@informatik.uni-mannheim.de}, \\ 
  \textsuperscript{2}{\tt \{iv250,ep490,alk23\}@cam.ac.uk}
}

\date{}

\begin{document}
\maketitle
\begin{abstract}
Unsupervised pretraining models have been shown to facilitate a wide range of downstream NLP applications. These models, however, retain some of the limitations of traditional static word embeddings. In particular, they encode only the distributional knowledge available in raw text corpora, incorporated through language modeling objectives. In this work, we complement such distributional knowledge with external lexical knowledge, that is, we integrate the discrete knowledge on word-level semantic similarity into pretraining. To this end, we generalize the standard BERT model to a multi-task learning setting where we couple BERT's masked language modeling and next sentence prediction objectives with an auxiliary task of binary word relation classification. Our experiments suggest that our ``Lexically Informed'' BERT (LIBERT), specialized for the word-level semantic similarity, yields better performance than the lexically blind ``vanilla'' BERT on several language understanding tasks. Concretely, LIBERT outperforms BERT in 9 out of 10 tasks of the GLUE benchmark and is on a par with BERT in the remaining one. Moreover, we show consistent gains on 3 benchmarks for lexical simplification, a task where knowledge about word-level semantic similarity is paramount. 
\end{abstract}

\section{Introduction}

Unsupervised pretraining models, such as GPT and GPT-2 \citep{Radford2018ImprovingLU,radford2019language}, ELMo \citep{peters-etal-2018-deep}, and BERT \citep{devlin2018bert} yield state-of-the-art performance on a wide range of natural language processing tasks. All these models rely on language modeling (LM) objectives that exploit the knowledge encoded in large text corpora. 
%
%However, they rely on the distributional signal only, coming from large human-generated corpora, which they incorporate in terms of language modeling objectives. 
%
BERT \cite{devlin2018bert}, as one of the current state-of-the-art models, is pretrained on a joint objective consisting of two parts: (1) masked language modeling (MLM), and (2) next sentence prediction (NSP). Through both of these objectives, BERT still consumes only the distributional knowledge encoded by word co-occurrences. 

While several concurrent research threads are focused on making BERT optimization more robust~\citep{liu2019roberta} or on imprinting external world knowledge on its representations~\citep[\textit{inter alia}]{sun2019ernie,zhang2019ernie,sun2019ernie2,liu2019k,peters2019knowledge}, no study yet has been dedicated to mitigating a severe limitation that contextualized representations and unsupervised pretraining inherited from static embeddings: every model that relies on distributional patterns has a tendency to conflate together pure lexical semantic similarity with broad topic relatedness \cite{Schwartz:2015conll,Mrksic:2017tacl}.

In the past, a plethora of models have been proposed for injecting linguistic constraints (i.e., lexical knowledge) from external resources to static word embeddings \cite[\textit{inter alia}]{Faruqui:2015naacl,Wieting:2015tacl,Mrksic:2017tacl, ponti2018adversarial} in order to emphasize a particular lexical relation in a \textit{specialized} embedding space. For instance, lexically informed word vectors specialized for pure semantic similarity result in substantial gains in a number of downstream tasks where such similarity plays an important role, e.g., in dialog state tracking \cite{Mrksic:2017tacl,Ren:2018emnlp} or for lexical simplification \cite{glavavs2018explicit,ponti2019cross}. 
%Likewise, specializing for lexical entailment \cite{vulic2018specialising} benefits taxonomy induction \cite{Nguyen:2017emnlp,nickel2018learning}. 
Existing specialization methods are, however, not directly applicable to unsupervised pretraining models because they are either (1) tied to a particular training objective of a static word embedding model, or (2) predicated on the existence of an embedding space in which pairwise distances can be modified.

In this work, we hypothesize that supplementing unsupervised LM-based pretraining with clean lexical information from structured external resources may also lead to improved performance in language understanding tasks. We propose a novel method to inject linguistic constraints, available from lexico-semantic resources like WordNet~\cite{Miller:1995:WLD:219717.219748} and BabelNet~\cite{navigli2012babelnet}, into unsupervised pretraining models, and steer them towards capturing word-level semantic similarity. To train Lexically Informed BERT (LIBERT), 
we (1) feed semantic similarity constraints to BERT as additional training instances and (2) predict lexico-semantic relations from the constraint embeddings produced by BERT's encoder \cite{vaswani2017attention}. In other words, LIBERT adds lexical relation classification (LRC) as the third pretraining task to BERT's multi-task learning framework.        

We compare LIBERT to a lexically blind ``vanilla'' BERT on the GLUE benchmark \citep{wang-etal-2018-glue} and report their performance on corresponding development and test portions. LIBERT yields performance gains over BERT on 9/10 GLUE tasks (and is on a par with BERT on the remaining one), with especially wide margins on tasks involving complex or rare linguistic structures such as Diagnostic Natural Language Inference and Linguistic Acceptability. Moreover, we assess the robustness and effectiveness of LIBERT on 3 different datasets for lexical simplification (LS), a task proven to benefit from word-level similarity specialization \cite{ponti2019cross}. We report LS improvements of up to 8.2\% when using LIBERT in lieu of BERT. For direct comparability, we train both LIBERT and BERT from scratch, and monitor the gains from specialization across iterations. Interestingly, these do not vanish over time, which seems to suggest that our specialization approach is suitable also for models trained on massive amounts of raw text data.
\section{Related Work}

\subsection{Specialization for Semantic Similarity}
\label{ss:rwspec}

The conflation of disparate lexico-semantic relations in \textit{static} word representations is an extensively researched problem. For instance, clearly discerning between true semantic similarity and broader conceptual relatedness in static embeddings benefits a range of natural language understanding tasks such as dialog state tracking \cite{Mrksic:2017tacl}, text simplification \cite{glavavs2018explicit}, and spoken language understanding \cite{Kim:2016slt}. The most widespread solution relies on the use of specialization algorithms to enrich word embeddings with external lexical knowledge and steer them towards a desired lexical relation. 

\textit{Joint specialization} models \cite[\textit{inter alia}]{Yu:2014,Kiela:2015emnlp,Liu:EtAl:15,Osborne:16,Nguyen:2017emnlp} jointly train word embedding models from scratch and enforce the external constraints with an auxiliary objective. On the other hand, \textit{retrofitting} models are post-processors that fine-tune pretrained word embeddings by gauging pairwise distances according to the external constraints \cite{Faruqui:2015naacl,Wieting:2015tacl,Mrksic:2016naacl,Mrksic:2017tacl,Jo:2018extro}. 
%These models generally achieve better downstream results and are more general, since they are decoupled from the word embedding objective.

More recently, retrofitting models have been extended to specialize not only words found in the external constraints, but rather the entire embedding space. In \textit{explicit retrofitting} models \citep{glavavs2018explicit}, a (deep, non-linear) specialization function is directly learned from external constraints. \textit{Post-specialization} models \cite{Vulic:2018naaclpost,ponti2018adversarial,kamath2019specializing}, instead, propagate lexico-semantic information to unseen words by imitating the transformation undergone by seen words during the initial specialization. This family of models can also transfer specialization across languages \cite{glavavs2018explicit,ponti2019cross}.  

The goal of this work is to move beyond similarity-based specialization of static word embeddings only. We present a novel methodology for enriching unsupervised pretraining models such as BERT \cite{devlin2018bert} with readily available discrete lexico-semantic knowledge, and measure the benefits of such semantic specialization on similarity-oriented downstream applications.

\subsection{Injecting Knowledge into Unsupervised Pretraining Models}
Unsupervised pretraining models do retain some of the limitations of static word embeddings. First, they still conflate separate lexico-semantic relations, as they learn from distributional patterns. Second, they fail to fully capture the world knowledge necessary for human reasoning: masked language models struggle to recover knowledge base triples from raw texts \citep{petroni2019language}. Recent work has, for the most part, focused on mitigating the latter limitation by injecting structured world knowledge into unsupervised pretraining and contextualized representations.

% it is not straightforward to apply some of the specialization techniques devised for static embeddings directly
% predicated on the permutation of a static space 

In particular, these techniques fall into the following broad categories: i) \textit{masking} higher linguistic units of meanings, such as phrases or named entities, rather than individual WordPieces or BPE tokens \citep{sun2019ernie}; ii) including an \textit{auxiliary task} in the objective, such as denoising auto-encoding of entities aligned with text \citep{zhang2019ernie}, or continuous learning frameworks over a series of unsupervised or weakly supervised tasks (e.g.,\ capitalization prediction or sentence reordering) \citep{sun2019ernie2}; iii) \textit{hybridizing} texts and graphs. \citet{liu2019k} proposed a special attention mask and soft position embeddings to preserve their graph structure while preventing unwanted entity-word interactions. \citet{peters2019knowledge} fuse language modeling with an end-to-end entity linker, updating contextual word representations with word-to-entity attention.

As the main contributions of our work, we incorporate external lexico-semantic knowledge, rather than world knowledge, in order to rectify the first limitation, namely the distortions originating from the distributional signal. In fact, \citet{liu2019k} hybridized texts also with linguistic triples relating words to sememes (minimal semantic components); however, this incurs into the opposite effect of reinforcing the distributional signal based on co-occurrence. On the contrary, we propose a new technique to enable the model to distinguish between purely similar and broadly related words.
\section{Specializing for Word-Level Similarity}

\begin{figure}[t!]
    \centering
    \includegraphics[width=1.0\linewidth,trim=0.2cm 0cm 0.2cm 0cm]{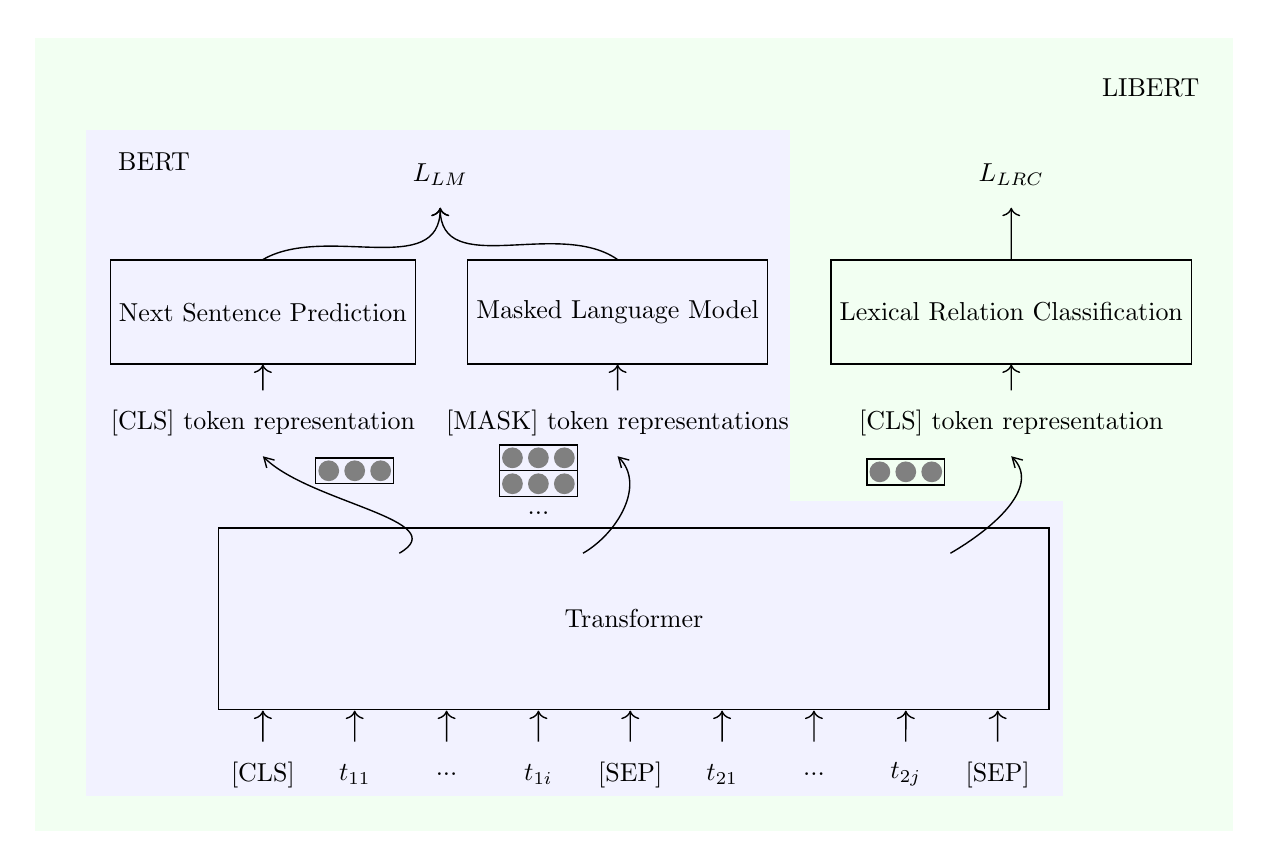}
    \vspace{-8mm}
    \caption{Architecture of LIBERT -- lexically-informed BERT specialized with semantic similarity constraints.}
    \label{fig:bert}
    \vspace{-3mm}
\end{figure}

LIBERT, illustrated in Figure~\ref{fig:bert}, is a \textit{joint} specialization model. It augments BERT's two pretraining tasks -- masked language modeling (1. MLM) and next sentence prediction (2. NSP) -- with an additional task of identifying (i.e., classifying) valid lexico-semantic relations from an external resource (3. LRC). LIBERT is first pretrained jointly on all three tasks. Similarly to BERT, after pretraining, LIBERT is fine-tuned on training datasets of downstream tasks. For completeness, we first briefly outline the base BERT model and then provide the details of our lexically informed augmentation.   

%applied in the pretraining stage. We next describe the generation of the additional input data required and our extensions regarding BERT's pretraining architecture and training setting.

\subsection{BERT: Transformer-Based Encoder}
The core of the BERT model is a multi-layer bidirectional Transformer \citep{vaswani2017attention}, pretrained using two objectives: (1) masked language modeling (MLM) and (2) next sentence prediction (NSP). MLM is a token-level prediction task, also referred to as \emph{Cloze} task \citep{taylor:cloze}: among the input data, a certain percentage of tokens is masked out and needs to be recovered. NSP operates on the sentence-level and can, therefore, be seen as a higher-level sequence modeling task that captures information across sentences. NSP predicts if two given sentences are adjacent in text (negative examples are created by randomly pairing sentences).

%Given two sentences,s1ands2, the task is to predict whethers2is a true suc-cessor ofs1. For the input data generation, 50% ofthe sentences are choosen randomly.

\subsection{LIBERT: Lexically-Informed (Specialized) Pretraining}

The base BERT model consumes only the distributional information. We aim to steer the model towards capturing true semantic similarity (as opposed to conceptual relatedness) by exposing it to clean external knowledge presented as the set of \textit{linguistic constraints} $C = \{(w_1, w_2)_i\}^N_{i = 1}$, i.e., pairs of words that stand in the desired relation (i.e., true semantic similarity) in some external lexico-semantic resource. Following the successful work on semantic specialization of static word embeddings (see \S\ref{ss:rwspec}), in this work we select pairs of synonyms (e.g., \textit{car} and \textit{automobile}) and direct hyponym-hypernym pairs (e.g., \textit{car} and \textit{vehicle}) as our semantic similarity constraints.\footnote{As the goal is to inform the BERT model on the relation of {true semantic similarity} between words \cite{Hill:2015cl}, according to prior work on static word embeddings \cite{Vulic:2018injecting}, the sets of both synonym pairs and direct hyponym-hypernym pairs are useful to boost the model's ability to capture true semantic similarity, which in turn has a positive effect on downstream language understanding applications.}

We transform the constraints from $C$ into a BERT-compatible input format and feed them as additional training examples for the model. The encoding of a constraint is then forwarded to the relation classifier, which predicts whether the input word pair represents a valid lexical relation.  

\vspace{1.6mm}
\noindent \textbf{From Linguistic Constraints to Training Instances.} 
We start from a set of linguistic constraints $C = \{(w_1, w_2)_i\}^N_{i = 1}$ and an auxiliary static word embedding space $\mathbf{X_{aux}} \in \mathbb{R}^d$. The space $\mathbf{X_{aux}}$ can be obtained via any standard static word embedding model such as Skip-Gram \cite{Mikolov2013distributed} or fastText \cite{Bojanowski:2017tacl} (used in this work). Each constraint $c = (w_1, w_2)$ corresponds to a true/positive relation of semantic similarity, and thus represents a \textit{positive} training example for the model. For each positive example $c$, we create corresponding negative examples following prior work on specialization of static embeddings \cite{Wieting:2015tacl,glavavs2018explicit,ponti2019cross}. We first group positive constraints from $C$ into mini-batches $B_p$ of size $k$. For each positive example $c = (w_1, w_2)$, we create two negatives $\hat{c}_1 = (\hat{w}_1, w_2)$ and $\hat{c}_2 = (w_1, \hat{w}_2)$ such that $\hat{w}_1$ is the word from batch $B_p$ (other than $w_1$) closest to $w_2$ and $\hat{w}_2$ the word (other than $w_2$) closest to $w_1$, respectively, in terms of the cosine similarity of their vectors in $\mathbf{X_{aux}}$. This way we create a batch $B_n$ of $2k$ negative training instances from a batch $B_p$ of $k$ positive training instances.

Next, we transform each instance (i.e., a pair of words) into a ``BERT-compatible'' format, i.e., into a sequence of WordPiece \cite{wu2016google} tokens.\footnote{We use the same 30K WordPiece vocabulary as \newcite{devlin2018bert}. Sharing WordPieces helps our word-level task as lexico-semantic relationships are similar for words composed of the same morphemes.} We split both $w_1$ and $w_2$ into WordPiece tokens, insert the special separator token (with a randomly initialized embedding) before and after the tokens of $w_2$ and prepend the whole sequence with BERT's sequence start token, as shown in this example for the constraint (\textit{mended}, \textit{regenerated}):\footnote{The sign \# denotes split WordPiece tokens.}

\vspace{1.5mm}    
\begin{center}
\setlength{\tabcolsep}{2pt}
\footnotesize{
\begin{tabular}{c c c c c c c c}
    \texttt{[CLS]} & \texttt{men} & \texttt{\#ded} & \texttt{[SEP]} & \texttt{reg} & \texttt{\#ener} & \texttt{\#ated} & \texttt{[SEP]} \\
    \texttt{0} & \texttt{0} & \texttt{0} & \texttt{0} & \texttt{1} & \texttt{1} & \texttt{1} & \texttt{1} 
\end{tabular}
}
\end{center}
\vspace{1.5mm}

\noindent As in the original work \cite{devlin2018bert}, we sum the WordPiece embedding of each token with the embeddings of the segment and position of the token. We assign the segment ID of \texttt{0} to the \texttt{[CLS]} token, all $w_1$ tokens, and the first \texttt{[SEP]} token; segment ID \texttt{1} is assigned to all tokens of $w_2$ and the final \texttt{[SEP]} token. 

%\noindent  \hspace{-2.5mm} men \hspace{-2.5mm} \#ded \hspace{-2.5mm} [SEP] \hspace{-2.5mm} reg \hspace{-2.5mm} \#ener \hspace{-2.5mm} \#ated \hspace{-2.5mm} [SEP]} \\
%   \texttt{0} \hspace{2.3em} \texttt{0} \texttt{0} \texttt{0} \texttt{1} \texttt{1} \texttt{1} \texttt{1}

\normalsize

\vspace{1.8mm}
\noindent \textbf{Lexical Relation Classifier.} 
Original BERT feeds Transformer-encoded token representations to two classifiers: MLM classifier (predicting the masked tokens), and the NSP classifier (predicting whether two sentences are adjacent). LIBERT introduces the third pretraining classifier: it predicts whether an encoded word pair represents a desired lexico-semantic relation (i.e., a positive example where two words stand in the relation of true semantic similarity -- synonyms or hypernym-hyponym pairs) or not. Let $\textbf{x}_\mathit{CLS} \in \mathbb{R}^H$ be the transformed vector representation of the sequence start token [CLS] that encodes the whole constraint ($w_1, w_2$). Our lexical relation predictor (LRC) is a simple softmax classifier formulated as follows:  
%
%It follows the original one of \citet{devlin2018bert} in most aspects, except that we extend the model with an additional classifier. To this end, we first pool the \texttt{[CLS]} token representation $C \in \mathbb{R}^H$ from the last layer of the transformer. Next we feed $C$ in a simple classification layer, defined as
%
\begin{equation}
    \hat{\mathbf{y}} = \mathrm{softmax}(\mathbf{x}_{\mathit{CLS}}\mathbf{W}_{\mathit{LRC}}^\top + \mathbf{b}_{\mathit{LRC}})\,,
\end{equation}
with $\mathbf{W}_{\mathit{LRC}} \in \mathbb{R}^{H \times 2}$ and $\mathbf{b}_{\mathit{LRC}} \in \mathbb{R}^{2}$ as the classifier's trainable parameters. Relation classification loss $L_{\mathit{LRC}}$ is then simply the negative log-likelihood over $k$ instances in the training batch:
\begin{equation}
    L_{\mathit{LRC}} = -\sum_{k}{\ln \mathbf{\hat{y}}_k \cdot \mathbf{y}_k}.
    \label{eq:lrc}
\end{equation}
\noindent where $\mathbf{y} \in \{[0, 1], [1, 0]\}$ is the true relation label for a word-pair training instance.

%with the aim to incorporate clean linguistic knowledge from the early training stages on.
%Given an input word pair $(w_a, w_b)_j$ from the set of positive and negative constraints $\mathbf{C}$ the task is to predict the associated binary similarity label $s_j \in S=\{\mathrm{true}, \mathrm{false}\}$, indicating whether the terms $w_a$ and $w_b$ are truely similar. The training objective is to minimize the binary corss-entropy loss defined as follows:

%{\small
%\begin{equation}
%L_{CP} = -(y \ln(p) + (1-y)\ln(1-p))\,. %\sum_o \sum^{2}_{k = 1}{y^{(k)}\cdot \ln \left(y'^{(k)}\right)}\,,
%\vspace{-1em}
%\end{equation}
%}
%The training is a multi-task learning setting in which we train the original objectives jointly, but we alternate with the linguistic constraint objective.
\section{Language Understanding Evaluation}
To isolate the effects of injecting linguistic knowledge into BERT, we train base BERT and LIBERT in the same setting: the only difference is that we additionally update the parameters of LIBERT's Transformer encoder based on the gradients of the LRC loss $L_{\mathit{LRC}}$ from Eq.~\eqref{eq:lrc}. In the first set of experiments, we probe the usefulness of injecting semantic similarity knowledge on the well-known suite of GLUE tasks \cite{wang-etal-2018-glue}, while we also present the results on lexical simplification, another task that has been shown to benefit from semantic similarity specialization \cite{glavavs2018explicit}, later in \S\ref{s:lexsimp}.

\subsection{Experimental Setup}
%and evaluate them on ten downsstream tasks: (1) The vanilla BERT model (\textsc{Base}), and the linguistically-informed extension (\textsc{Informed}).

%\vspace{1.6mm}
\noindent \textbf{Pretraining Data.}
We minimize BERT's original objective $L_\mathit{MLM} + L_\mathit{NSP}$ on training examples coming from English Wikipedia.\footnote{We acknowledge that training the models on larger corpora would likely lead to better absolute downstream scores; however, the main goal of this work is not to achieve state-of-the-art downstream performance, but to compare the base model against its lexically informed counterpart.} We obtain the set of constraints $C$ for the $L_{\mathit{LRC}}$ term from the body of previous work on semantic specialization of static word embeddings \cite{Zhang:2014emnlp,Vulic:2018naaclpost,ponti2018adversarial}. In particular, we collect 1,023,082 synonymy pairs from WordNet \cite{Miller:1995:WLD:219717.219748} and Roget's Thesaurus \cite{Kipfer:2009book} and 326,187 direct hyponym-hypernym pairs \cite{vulic2018specialising} from WordNet.\footnote{Note again that similar to work of \citet{Vulic:2018injecting}, both WordNet synonyms and direct hyponym-hypernym pairs are treated exactly the same: as positive examples for the relation of true semantic similarity.}

%\alert{Were are the actual linguistic constraints coming from? Are they purely coming from WordNet? \citep{Miller:1995:WLD:219717.219748}}. 

%\setlength{\tabcolsep}{5pt}

\vspace{1.8mm}
\noindent \textbf{Fine-Tuning (Downstream) Tasks.} We evaluate BERT and LIBERT on the the following tasks from the GLUE benchmark \citep{wang-etal-2018-glue}, where sizes of training, development, and test datasets for each task are provided in Table~\ref{tbl:glue}: 
\vspace{1.4mm}

\noindent \textbf{CoLA} \cite{warstadt2018neural}: Binary sentence classification, predicting if sentences from linguistic publications are grammatically acceptable;

%The \textbf{Corpus of Linguistic Acceptability (CoLA)} consists of example sentences from linguistic publications annotated for their grammaticality \citep{warstadt2018neural}. The task is a binary acceptability classification and the evaluation measure is Matthews Correlation Coefficient (MCC).

\vspace{1.3mm}
\noindent \textbf{SST-2} \cite{socher2013recursive}: Binary sentence classification, predicting sentiment (positive or negative) for movie review sentences;

\vspace{1.3mm}
\noindent \textbf{MRPC} \cite{dolan2005automatically}: Binary sentence-pair classification, predicting whether two sentences are mutual paraphrases; 
%The measures employed are $F1$-measure and accuracy. 

\vspace{1.3mm}
\noindent \textbf{STS-B} \cite{cer-etal-2017-semeval}: Sentence-pair regression task, predicting the degree of semantic similarity for a pair of sentences;

%and the prediction quality is evaluated by computing the Pearson and Spearman correlation coefficients between the prediction and the ground truth.

\vspace{1.3mm}
\noindent \textbf{QQP} \cite{chen2018quora}: Binary classification task, recognizing question paraphrases;
%The result is evaluated in terms of $F1$-measure and accuracy.

\vspace{1.3mm}
\noindent \textbf{MNLI} \cite{williams2018broad}: Ternary natural language inference (NLI) classification of sentence pairs. Two test sets are given: a matched version (MNLI-m) in which the test domains match with training data domains, and a mismatched version (MNLI-mm) with different test domains;
%The performance of the models is reported in accuracy. 

\vspace{1.3mm}
\noindent \textbf{QNLI}: A binary classification version of the Stanford Q\&A dataset \citep{rajpurkar-etal-2016-squad}; 

\vspace{1.3mm}
\noindent \textbf{RTE} \cite{bentivogli2009fifth}: Another NLI dataset, ternary entailment classification for sentence pairs; 

\vspace{1.3mm}
\noindent \textbf{AX} \cite{wang-etal-2018-glue}: A small, manually curated NLI dataset (i.e., a ternary classification task), with examples encompassing different linguistic phenomena relevant for entailment.\footnote{Following \newcite{devlin2018bert}, we do not evaluate on the Winograd NLI (WNLI), given its well-documented issues.}
%\vspace{1.4mm}

%\noindent 

\vspace{1.8mm}
\noindent \textbf{Training and Evaluation.}
We train both BERT and LIBERT from scratch, with the configuration of the BERT$_{BASE}$ model \cite{devlin2018bert}: $L=12$ transformer layers with the hidden state size of $H=768$, and $A=12$ self-attention heads. We train in batches of $k = 16$ instances;\footnote{Due to hardware restrictions, we train in smaller batches than in the the original work \cite{devlin2018bert} ($k = 256$). This means that for the same number of update steps, our models will have observed less training data than the original BERT model of \newcite{devlin2018bert}.}
%\footnote{Due to hardware restrictions, we train in smaller batches that are half the size of the training batches from the original work \cite{devlin2018bert} ($k = 32$). This means that for the same number of update steps, our models will have seen half of the amount of the original BERT model of \newcite{devlin2018bert}.}
the input sequence length is $128$. The learning rate for both models is $2 \cdot 10^{-5}$ with a warm-up over the first $1,000$ training steps. Other hyperparameters are set to the values reported by \newcite{devlin2018bert}. 

LIBERT combines BERT's MLM and NSP objectives with our LRC objective in a multi-task learning setup. We update its parameters in a balanced alternating regime: (1) we first minimize BERT's $L_\mathit{MLM} + L_\mathit{NSP}$ objective on one batch of masked sentence pairs and then (2) minimize the LRC objective $L_\mathit{LRC}$ on one batch of training instances created from linguistic constraints. 

\begin{table*}[!t]
\centering
\def\arraystretch{0.95}
\small{
{%\fontsize{8pt}{8pt}\selectfont
\begin{tabularx}{\linewidth}{l X X X X X X X X X X}
\toprule
  & CoLA & SST-2 & MRPC & STS-B & QQP & MNLI-m & MNLI-mm & QNLI & RTE & AX\\
  \midrule
  \# Train & 8,551 & 67,349 & 3,668 & 5,749 & 363,870 & 392,702 & 392,702 & 104,743 & 2,490 & -- \\
  \# Dev & 1,042 & 872 & 408 & 1,501 & 40,431 & 9,815 & 9,832 & 5,463 & 278 & --\\
  \# Test & 1,063 & 1,821 & 1,725 & 1,379 & 390,964 & 9,796 & 9,847 & 5,463 & 3,000 & 1,104\\
%  \midrule
%  \# Total \\
\bottomrule
\end{tabularx}
}}
\vspace{-0.5mm}
\caption{dataset sizes for tasks in the GLUE benchmark \citep{wang-etal-2018-glue}.}
\label{tbl:glue}
\vspace{-0.5mm}
\end{table*}
\newcolumntype{g}{>{\columncolor{white}}l}
\setlength{\tabcolsep}{2pt}
\begin{table*}[!t]
\def\arraystretch{0.97}
\centering
%\small{
{\fontsize{8pt}{8pt}\selectfont
\begin{tabularx}{\linewidth}{l l l X X X X X X X X X X}
\toprule
&  & & CoLA & SST-2 & MRPC & STS-B & QQP & MNLI-m & MNLI-mm & QNLI & RTE & AX\\

&  & & MCC & Acc & F1/Acc & Pears & F1/Acc & Acc & Acc & Acc & Acc & MCC\\
\midrule
&    \multirow{3}{3em}{Dev} & BERT & 29.4 & 88.7 & 87.1/81.6 & 86.4 & 85.9/89.5 &  78.2 & \textbf{78.8} & 86.2 & 63.9 & -- \\
&  & LIBERT & \textbf{35.3} & \textbf{89.9} & \textbf{87.9/82.6} & \textbf{87.2} & \textbf{86.3/89.8} & \textbf{78.5} & 78.7 & \textbf{86.5} & \textbf{65.3} & --\\
\rowcolor{Gray}
&  & $\Delta$ & +5.9 & +1.2 & +0.8/+1.0 & +0.8 & +0.4/+0.3 & +0.3 & -0.1 & +0.3 & +1.4 & -- \\
%  \midrule
\multirow{-2}{3em}{1M} &  \multirow{3}{3em}{Test} & BERT & 21.5 & 87.9 & 84.8/78.8 & \textbf{80.8} & 68.6/87.9 & 78.2 & \textbf{77.6} & 85.8 & 61.3 & 26.8 \\
&  & LIBERT & \textbf{31.4} & \textbf{89.6} & \textbf{86.1/80.4} & 80.5 & \textbf{69.0/88.1} & \textbf{78.4} & 77.4 & \textbf{86.2} & \textbf{62.6} & \textbf{32.8} \\
\rowcolor{Gray}
&  & $\Delta$ & +9.9 & +1.7 & +1.3/+1.6 & -0.3 & +0.4/+0.2& +0.2 & -0.2& +0.4 & +1.3 & +6.0\\
%%% this is for the 2M steps model
\midrule
& \multirow{3}{3em}{Dev} & BERT & 30.0 & 88.5 & 86.4/81.1 & 87.0 & 86.3/89.8 & 78.8 & 79.3 & 86.6 & 64.3 & -- \\
&   & LIBERT & \textbf{37.2} & \textbf{89.3} & \textbf{88.7}/\textbf{84.1} & \textbf{88.3} & \textbf{86.5/90.0} & \textbf{79.6} & \textbf{80.0} & \textbf{87.7} & \textbf{66.4} & --\\
\rowcolor{Gray}
& & $\Delta$ &  +7.2 & +0.8 & +2.3/+3.0 & +1.3 & +0.2/+0.2 & +0.8 & +0.7 & +1.1 & +2.1 & --\\
%  \midrule
\multirow{-2}{3em}{2M} &\multirow{3}{3em}{Test} & BERT & 28.8 & 89.7 & 84.9/79.1 & 81.1 & 69.0/88.0 & 78.6 & 78.1 & \textbf{87.2} & 63.4 & 30.8\\
& & LIBERT & \textbf{35.3} & \textbf{90.8} & \textbf{86.6/81.7} & \textbf{82.6} & \textbf{69.3/88.2} & \textbf{79.8} & \textbf{78.8} & \textbf{87.2} & \textbf{63.6} & \textbf{33.3}\\
\rowcolor{Gray}
& & $\Delta$ & +6.5 & +1.1 & +1.7/+2.6 & +1.5 &+0.3/+0.2 & +1.2 & +0.7 & +0.0 & +0.2 & +2.5\\
\bottomrule
\end{tabularx}
}%}
\vspace{-0.5mm}
\caption{Results on 10 GLUE tasks after 1M and 2M MLM+NSP steps with BERT and LIBERT.}
\label{tbl:results}
\vspace{-1.5mm}
\end{table*}

\begin{figure*}[!t]
    \centering
    \begin{subfigure}[t]{0.48\linewidth}
        \centering
        \includegraphics[width=0.99\linewidth]{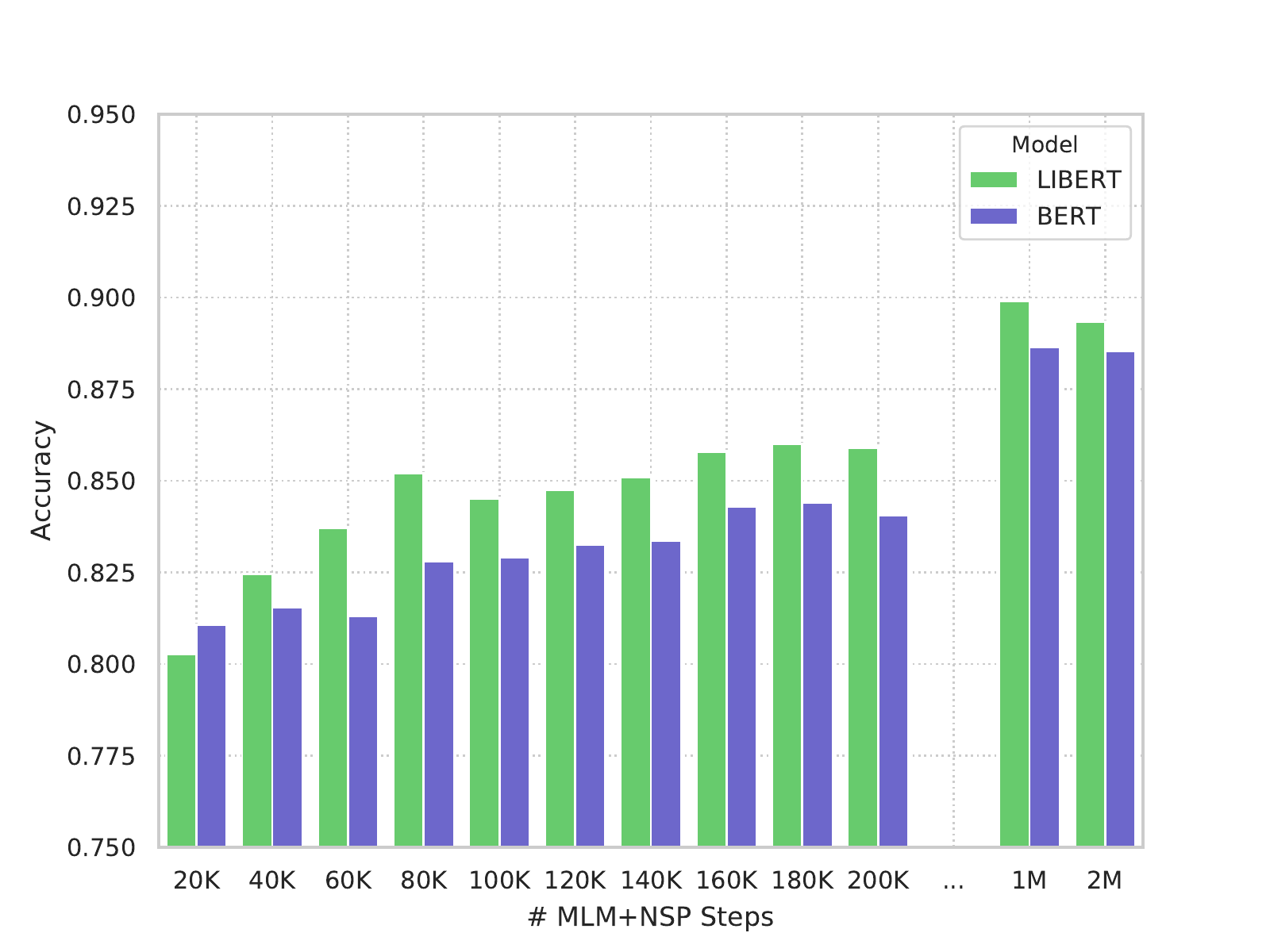}
        %\vspace{-0.7em}
        \caption{SST-2}
        \label{fig:sst2}
    \end{subfigure}
    \begin{subfigure}[t]{0.48\textwidth}
        \centering
        \includegraphics[width=0.99\linewidth]{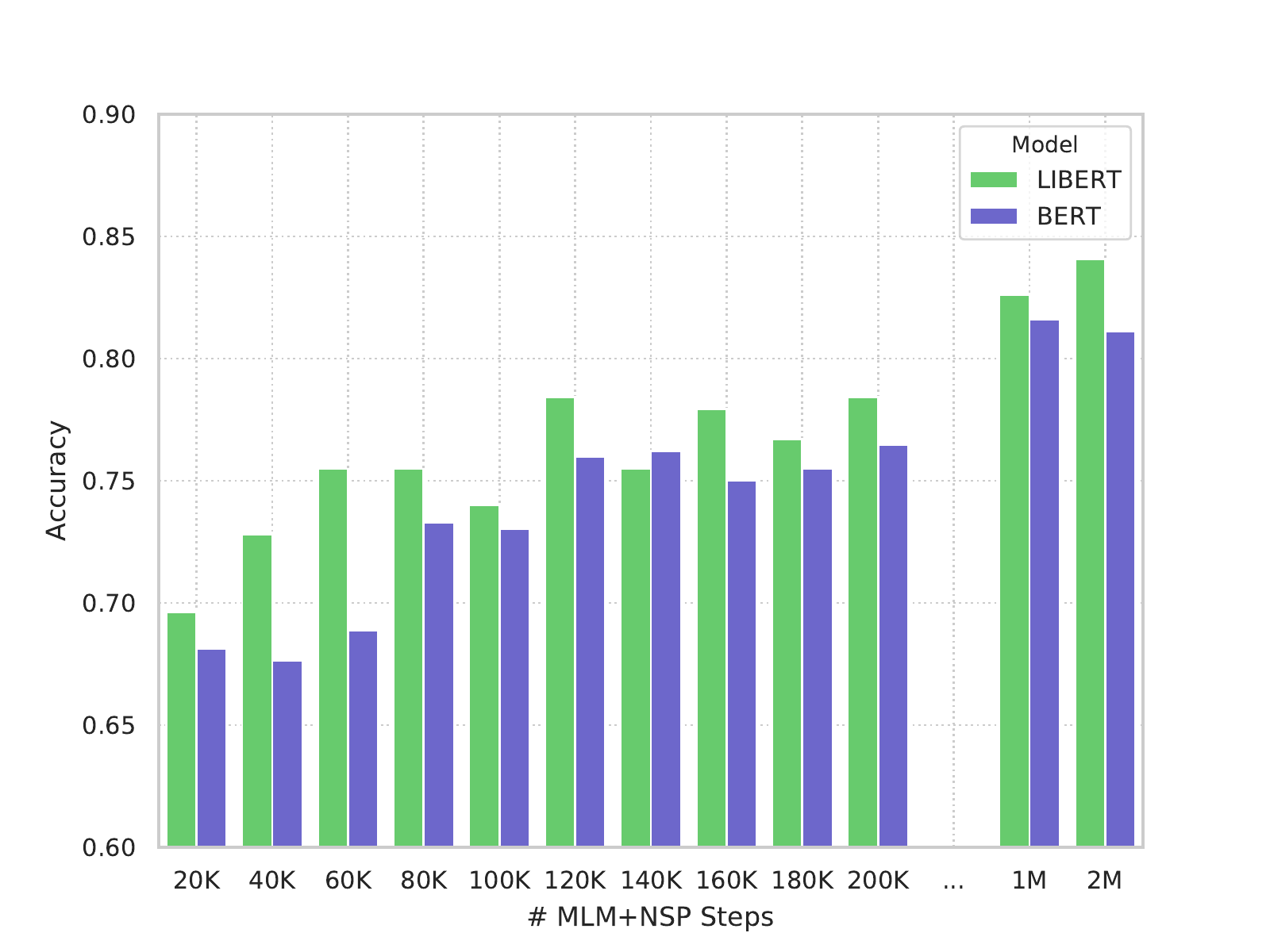}
        %\vspace{-0.7em}
        \caption{MRPC}
        \label{fig:mrpc}
    \end{subfigure}

    \vspace{-0.5mm}
    \caption{Accuracy over time for BERT and LIBERT on (a) SST-2 and (b) MRPC on the corresponding dev sets.}
    \vspace{-0.5mm}
\label{fig:similar}
\end{figure*}

\begin{table*}[th]
    \centering
    \small{
\begin{tabularx}{\linewidth}{XX|X|XXXX|XXXXXX}
\toprule
& & & \multicolumn{4}{c|}{Coarse-grained} & \multicolumn{6}{c}{Fine-grained} \\
& Model &  All & LS & PAS & Lo & KCS & LE & MN & Fa & Re & NE & Qu \\
\hline
\multirow{3}{*}{1M} & BERT & 26.8 & 24.5 & 38.8 & 19.6 & 12.8 & 17.5 & 29.3 & 04.9 &  22.5 & 15.6 & \textbf{57.2} \\
& LIBERT & \textbf{32.8} & \textbf{35.2} & \textbf{39.7} & \textbf{25.3} & \textbf{19.4} & \textbf{28.5} & \textbf{51.4} & \textbf{18.7} & \textbf{59.2} & \textbf{18.0} & {56.9} \\
\rowcolor{Gray} &
  $\Delta$ & 6.0 & 10.7 & 0.9 & 5.7 & 6.6 & 11.0 & 22.2 & 13.8 & 36.7 & 2.4 & -0.3\\
 \hline
 \multirow{3}{*}{2M} & BERT & 30.8 & 31.3 & 40.0 & 21.7 & \textbf{19.7} & 21.2 & 51.3 & 09.1 & 59.2 & \textbf{21.0} & 60.5 \\
 & LIBERT & \textbf{33.3} & \textbf{40.6} & 39.9 & \textbf{24.5} & 18.3 & \textbf{33.2} & \textbf{72.0} & \textbf{21.0} & 59.2 & 18.3 & \textbf{68.4} \\
 \rowcolor{Gray} &
  $\Delta$ & 2.5 & 9.3 & -0.1 & 2.8 & -1.4 & 12.0 & 20.7 & 11.9 & 0.0 & -2.7 & 7.9 \\
\bottomrule
\end{tabularx}
}%
    \caption{Linguistic analysis on the Diagnostic dataset. The scores are $R_3$ coefficients between gold and predicted labels, scaled by 100, for sentences containing linguistic phenomena of interest. We report all the coarse-grained categories: \textit{Lexical Semantics} (\textbf{LS}), \textit{Predicate-Argument Structure} (\textbf{PAS}), \textit{Logic} (\textbf{Lo}), and \textit{Knowledge and Common Sense} (\textbf{KCS}). Moreover, we report fine-grained categories for Lexical Semantics: \textit{Lexical Entailment} (\textbf{LE}), \textit{Morphological Negation} (\textbf{MN}), \textit{Factivity} (\textbf{Fa}), \textit{Redundancy} (\textbf{Re}), \textit{Named Entities} (\textbf{NE}), and \textit{Quantifiers} (\textbf{Qu}).}
    \label{tab:diagnostic}
\end{table*}

%\footnote{In the original work, BERT was trained in batches of $k = 32$ instances. 
%For the same number of batches, our models will have seen half of the amount of data compared to the original work.} 

During fine-tuning, for each task, we independently find the optimal hyperparameter configurations of the downstream classifiers for the pretrained BERT and LIBERT: this implies that it is valid to compare their performances on the downstream development sets. Finally, we evaluate fine-tuned BERT and LIBERT on all 10 test sets.

%% we continue training the Transformer components on the training portions of downstream tasks
%% -> This is the established fact.
%% albeit optimistic in terms of absolute scores,

\subsection{Results and Discussion}

\textbf{Main Results.} The main results are summarized in Table~\ref{tbl:results}: we report both dev set and test set performance. After 1M MLM+NSP steps, LIBERT outperforms BERT on 8/9 tasks (dev) and 8/10 tasks (test). After 2M MLM+NSP steps, LIBERT is superior in 9/9 tasks (dev) and 9/10 tasks (test). For the test set of the tenth task (QNLI), LIBERT is on a par with BERT. While large gains are reported on CoLA, AX, and visible gains appear on SST-2 and MRPC, it is encouraging to see that slight and consistent gains are observed on almost all other tasks. These results suggest that available external lexical knowledge can be used to supplement unsupervised pretraining models with useful information which cannot be fully captured solely through large text data and their distributional signal. The results indicate that LIBERT, our lexically informed multi-task method, successfully blends such curated linguistic knowledge with distributional learning signals. It also further validates intuitions from relevant work on specializing static word embeddings \cite{Wieting:2015tacl,Mrksic:2017tacl} that steering distributional models towards capturing true semantic similarity (as also done here) has a positive impact on language understanding applications in general.

\vspace{1.8mm}
\noindent \textbf{Fine-grained Analysis.}
To better understand how lexical information corroborates the model predictions, we perform a fine-grained analysis on the Diagnostic dataset \cite{wang-etal-2018-glue}, measuring the performance of LIBERT on specific subsets of sentences annotated for the linguistic phenomena they contain. We report the results in Table~\ref{tab:diagnostic}. As expected, \textit{Lexical Semantics} is the category of phenomena that benefits the most (+43.7\% for 1M iterations, +29.7\% for 2M), but with significant gains also in phenomena related to \textit{Logic} (+29.1\% for 1M and +29.1\% for 2M) and \textit{Knowledge \& Common Sense} (+51.7\% for 1M). Interestingly, these results seem to suggest that knowledge about semantic similarity and lexical relations also partially encompasses factual knowledge about the world. 

By inspecting even finer-grained phenomena related to \textit{Lexical Semantics}, LIBERT outdistances its baseline by a large margin in: i) \textit{Lexical Entailment} (+62.9\% for 1M, +56.6\% for 2M), as expected from the guidance of hypernym-hyponym pairs; ii) \textit{Morphological Negation} (+75.8\% for 1M, +40.4\% for 2M). Crucially, the lower performance of BERT cannot be explained by the low frequency of morphologically derived words (prevented by the WordPiece tokenization), but exactly because of the distributional bias. iii) \textit{Factivity} (+281.7\% for 1M, +130.8\% for 2M), which is a lexical entailment between a clause and the entire sentence it is embedded in. Since it depends on specific lexical triggers (usually verbs or adverbs), it is clear that lexico-semantic knowledge better characterizes the trigger meanings. The improvement margin for \textit{Redundancy} and \textit{Quantifiers} fluctuate across different amounts of iterations, hence no conclusions can be drawn from the current evidence.

\vspace{1.8mm}
\noindent \textbf{Performance over Time.} 
Further, an analysis of performance over time (in terms of MLM+NSP training steps for BERT and LIBERT) for one single-sentence task (SST-2) and one sentence-pair classification task (MRPC) is reported in Figures~\ref{fig:sst2}-\ref{fig:mrpc}. The scores clearly suggest that the impact of external knowledge does not vanish over time: the gains with the lexically-informed LIBERT persist at different time steps. This finding again indicates the complementarity of useful signals coded in large text data versus lexical resources \cite{Faruqui:2016thesis,Mrksic:2017tacl}, which should be investigated more in future work.
\section{Similarity-Oriented Downstream Evaluation: Lexical Simplification}
\label{s:lexsimp}
%We report the experimental setup and results in the similarity oriented downstream evaluation.
%\subsection{Lexical Simplification}
%
%
%
%
\setlength{\tabcolsep}{3.7pt}
\begin{table*}[!t]
\centering
\small{
%{\fontsize{8pt}{8pt}\selectfont
\begin{tabular}{c l | c c c c c c c c c | c c c}%{\linewidth}{l c c c c c c c c c}
\toprule
& & \multicolumn{9}{c}{Candidate Generation} & \multicolumn{3}{c}{Full Simplification Pipeline}\\
& & \multicolumn{3}{c}{BenchLS} & \multicolumn{3}{c}{LexMTurk} & \multicolumn{3}{c}{NNSeval} & BenchLS & LexMTurk & NNSeval \\
\# Steps &  & P & R & F1 & P & R & F1 & P & R & F1 & A & A & A\\
\midrule
\multirow{3}{3em}{1M} & BERT & .2167 & .1765 & .1945 & .3043 & .1420 & .1937 & .1499 & .1200 & .1333 & .3854 & .5260 & .2469\\
& LIBERT & \textbf{.2348}	& \textbf{.1912} & \textbf{.2108} & \textbf{.3253} & \textbf{.1518}	& \textbf{.2072} & \textbf{.1646}	& \textbf{.1318} & \textbf{.1464} & \textbf{.4338} &	\textbf{.6080} & \textbf{.2678}\\
  \rowcolor{Gray}
& $\Delta$  & .0181 & .0147 & .0163 & .0210 & .0098 & .0135 & .0147 & .0118 & .0131 & .0484	& .0820	& .0209\\
\midrule
\multirow{3}{3em}{2M} & BERT & .2408 & .1960 & .2161 & .3267 & .1524 & .2079 & .1583 & .1267 & .1408 & .4241 & .5920	& .2594\\
& LIBERT  & \textbf{.2766} &\textbf{.2252} & \textbf{.2483} & \textbf{.3700} & \textbf{.1727} & \textbf{.2354} & \textbf{.1925} & \textbf{.1541} & \textbf{.1712} & \textbf{.4887} & \textbf{.6540} & \textbf{.2803} \\
  \rowcolor{Gray}
& $\Delta$  & .0358 & .0292 & .0322 & .0433 & .0203 & .0275 & .0342 & .0274 & .0304 & .0646	& .0620 & .0209\\
\bottomrule
\end{tabular}
}%}
%\vspace{-1.5mm}
\caption{Results on the lexical simplification candidate generation task and for the full pipeline on three datasets: BenchLS, LexMTurk, and NNSeval. For each dataset we report the performance after 1M and 2M MLM+NSP steps (\# Steps) with BERT and LIBERT in terms of Precision (P), Recall (R) and F1-measure (F1) for candidate generation and accuracy (A) for the full pipeline.}
\label{tbl:results_simpl}
%\vspace{-3.5mm}
\end{table*}
%
%
%
%\paragraph{Experimental Setup.}
% Task description

\textbf{Task Description.} The goal of lexical simplification is to replace a target word $w$ in a context sentence $S$ with simpler alternatives of equivalent meaning. Generally, the task can be divided into two main parts: (1) generation of substitute candidates, and (2) candidate ranking, in which the simplest candidate is selected \citep{paetzold2017survey}. 
% Model
Unsupervised approaches to candidate generation seem to be predominant lately \citep[e.g.,][]{glavas-stajner-2015-simplifying,ponti2019cross}. In this task, discerning between pure semantic similarity and broad topical relatedness (as well as from other lexical relations such as antonymy) is crucial. Consider the example: ``\textit{Einstein unlocked the door to the atomic age,}'' where \textit{unlocked} is the target word. In this context, the model should avoid confusion both with related words (e.g.\ \textit{repaired}) and opposite words (e.g.\ \textit{closed}) that fit in context but alter the original meaning.

\vspace{1.8mm}
\noindent \textbf{Experimental Setup.}
In order to evaluate the simplification capabilities of LIBERT versus BERT, we adopt a standard BERT-based approach to lexical simplification \citep{qiang2019BERTLS}, dubbed BERT-LS. It exploits the BERT MLM pretraining task objective for candidate generation.
% Candidate generation
Given the complex word $w$ and a context sentence $S$, we mask $w$ in a new sequence $S'$. Next, we concatenate $S$ and $S'$ as a sentence pair and create the BERT-style input by running WordPiece tokenization on the sentences, adding the \texttt{[CLS]} and \texttt{[SEP]} tokens before, in-between, and after the sequence, and setting segment IDs accordingly. We then feed the input either to BERT or LIBERT, and obtain the probability distribution over the vocabulary outputted by the MLM predictor based on the masked token $p(\cdot|S,S'\backslash\{w\} )$. Based on this, we select the candidates as the top $k$ words according to their probabilities, excluding morphological variations of the masked word.
% Substitution ranking --> which level of detail is needed? Also, we could put the algorithm, but I think as one of two main downstreams this level of detail is enough

For the substitution ranking component, we also follow \citet{qiang2019BERTLS}. Given the set of candidate tokens $C$, we compute for each $c_i$ in $C$ a set of features: (1) BERT prediction probability, (2) loss of the likelihood of the whole sequence according to the MLM when choosing $c_i$ instead of $w$, (3) semantic similarity between the fastText vectors~\citep{Bojanowski:2017tacl} of the original word $w$ and the candidate $c_i$, and (4) word frequency of $c_i$ in the top $12$ million texts of Wikipedia and in the Children's Book Test corpus.\footnote{A detailed description of these features can be found in the original work.} Based on the individual features, we next rank the candidates in $C$ and consequently, obtain a set of ranks for each $c_i$. The best candidate is chosen according to its average rank across all features.
In our experiments, we fix the number of candidates $k$ to $6$.
% datasets

\vspace{1.8mm}
\noindent \textbf{Evaluation Data.}
We run the evaluation on three standard datasets for lexical simplification:

\vspace{1.3mm}
\noindent (1) LexMTurk \citep{horn-etal-2014-learning}. The dataset consists of 500 English instances, which are collected from Wikipedia. The complex word and the simpler substitutions were annotated by $50$ crowd workers on Amazon Mechanical Turk.

\vspace{1.3mm}
\noindent (2) BenchLS \citep{paetzold-specia-2016-benchmarking} is a merge of LexMTurk and LSeval~ \citep{de2010text} containing 929 sentences. The latter dataset focuses on text simplification for children. The authors of BenchLS applied additional corrections over the instances of the two datasets.

\vspace{1.3mm}
\noindent (3) NNSeval \citep{paetzold2017survey} is an English dataset focused on text simplification for non-native speakers and consists in total of 239 instances. Similar to BenchLS, the dataset is based on LexMTurk, but filtered for a) instances that contain a complex target word for non-native speakers, and b) simplification candidates that were found to be non-complex by non-native speakers.

\vspace{1.3mm}
We report the scores on all three datasets in terms of Precision, Recall and F1 for the candidate generation sub-task, and in terms of the standard lexical simplification metric of \textit{accurracy} (A) \cite{horn-etal-2014-learning,glavas-stajner-2015-simplifying} for the full simplification pipeline. This metric computes the number of correct simplifications (i.e., when the replacement made by the system is found in the list of gold standard replacements) divided by the total number of target complex words.

% Do we have to describe what accuracy is in the context of LS?
\vspace{1.3mm}
\noindent \textbf{Results and Discussion.} The results for BERT and LIBERT for the simplification candidate generation task and for the full pipeline evaluation are provided in Table~\ref{tbl:results_simpl}. We report the performance of both models after 1M and 2M MLM+NSP pretraining steps. We observe that LIBERT consistently outperforms BERT by at least 0.9 percentage points across all evaluation setups, measures, and for all three evaluation sets. Same as in GLUE evaluation, the gains do not vanish as we train both models for a longer period of time (i.e., compare the differences between the two models after 1M vs.\ 2M training steps). On the contrary, for the candidate generation task, the gains of LIBERT over BERT are even higher after 2M steps. The gains achieved by LIBERT are also visible in the full simplification pipeline: e.g., on LexMTurk, replacing BERT with LIBERT yields a gain of 8.2 percentage points. In sum, these results confirm the importance of similarity specialization for a similarity-oriented downstream task such as lexical simplification.

\section{Conclusion}

We have presented LIBERT, a lexically informed extension of the state-of-the-art unsupervised pretraining model BERT. Our model is based on a multi-task framework that allows us to steer (i.e., specialize) the purely distributional BERT model to accentuate a lexico-semantic relation of true semantic similarity (as opposed to broader semantic relatedness). The framework combines standard BERT objectives with a third objective formulated as a relation classification task. The gains stemming from such explicit injection of lexical knowledge into pretraining were observed for 9 out of 10 language understanding tasks from the GLUE benchmark, as well as for 3 lexical simplification benchmarks. These results suggest that complementing distributional information with lexical knowledge is beneficial for unsupervised pretraining models. 

In the future, we will work on more sophisticated specialization methods, and we will investigate methods to encode the knowledge on asymmetric relations such as meronymy and lexical entailment. Finally, we will port this new framework to other languages and to resource-poor scenarios. We will release the code at: \url{[URL]}

\bibliography{references}
\bibliographystyle{acl_natbib}

%\appendix

%\section{Appendices}
%\label{sec:appendix}

%\section{Supplemental Material}
%\label{sec:supplemental}

\end{document}